# Enhancing Classification of Streaming Data with Image Distillation

Rwad Khatib, Yehudit Aperstein

Intelligent Systems, Afeka Academic College of Engineering, Tel Aviv

**Abstract.** This study tackles the challenge of efficiently classifying streaming data in environments with limited memory and computational resources. It delves into the application of data distillation as an innovative approach to improve the precision of streaming image data classification. By focusing on distilling essential features from data streams, our method aims to minimize computational demands while preserving crucial information for accurate classification. Our investigation compares this approach against traditional algorithms like Hoeffding Trees and Adaptive Random Forest, adapted through embeddings for image data. The Distillation Based Classification (DBC) demonstrated superior performance, achieving a 73.1% accuracy rate, surpassing both traditional methods and Reservoir Sampling Based Classification (RBC) technique. This marks a significant advancement in streaming data classification, showcasing the effectiveness of our method in processing complex data streams and setting a new standard for accuracy and efficiency.

## 1   Introduction

In the rapidly advancing field of artificial intelligence, the classification of streaming data represents a significant frontier, especially given its relevance to real-world applications ranging from financial fraud detection to healthcare monitoring and social media analytics. The ability to process and classify data in real time is paramount, as it supports immediate decision-making and insights into dynamic processes. However, the nature of streaming data—characterized by its high volume, velocity, and variety—presents unique challenges, particularly in environments constrained by limited computational resources and memory.

Classifying streaming data requires models that are efficient, accurate, and adaptable to concept drift. While Hoeffding Trees (HT) and Adaptive Random Forest (ARF) are key for incremental learning and handling real-time data, they face issues like managing concept drift, performance limitations, and computational demands. These drawbacks underscore a significant gap in current streaming data classification methods, especially in resource-constrained or complex scenarios.

This research aims to address these challenges by exploring the potential of data distillation techniques to enhance the classification of streaming data. Data distillation, which simplifies data representations while preserving essential information, promises a novel approach to improving model performance under continuous data flow conditions. By focusing on the efficacy of image distillation in streaming data classification, this work seeks to bridge the gap identified in existing methods, offering a solution that combines efficiency, accuracy, and adaptability.

The paper is structured as follows: Section 2 reviews related work and our contributions. Section 3 outlines our methodology. Results are presented in Section 4, with conclusions and future directions discussed in Section 5.

## 2    Previous Work

The implications of streaming data classification extend far beyond academic interest, finding practical applications in fields such as fraud detection, where real-time analysis is crucial for identifying and mitigating fraudulent transactions, and healthcare, where continuous monitoring of patient data can lead to timely interventions and improved patient outcomes.

Classification of streaming data, also referred to as stream learning or online learning, is a crucial area of research within machine learning and data mining. This domain focuses on developing models that incrementally learn from data streams continuously generated over time, evolving, and adapting to new patterns while maintaining previously learned knowledge. In addition, this field contrasts with traditional batch learning by focusing on models that adapt continuously as new data arrives, under constraints like limited memory and processing resources.

The framework for streaming data classification was laid out by early works that introduced the core concepts and initial algorithms. Gaber et al. [1] and Domingos et al. [2], set the stage for subsequent developments in streaming data classification. These works addressed the unique challenges of learning from continuous data streams, including concept drift, resource constraints, and the potentially infinite length of data streams.

In this area concept drift remains a central challenge, thoroughly review by Gama et al. [3], which has been extended by later studies like Webb et al. [4], who introduced advanced methods for detecting and adapting to concept drift in real time.

The constraints on memory and computation power in streaming settings have led to innovative solutions. Bifet and Gavalda[5] discussed adaptive windowing techniques to manage resource utilization effectively.

In algorithmic advances within streaming data analysis, both HT and ensemble strategies like ARF, introduced by Gomez et al. [6], have proven to be effective. These methods, by leveraging the collective strength of multiple learners for ARF and employing statistical confidence for decision-making in HT, enhance accuracy and adaptability in processing dynamic data streams. Oza and Russel [7] used bagging and boosting for online learning, which has been further refined by recent works that explore ensemble diversity and weighting strategies to improve performance in streaming contexts [8]. In addition, the exploration of unsupervised and semi-unsupervised learning models for streaming data has gained momentum, recognizing the potential for these methods to leverage unlabeled data effectively. This shift acknowledges the abundance of unlabeled data in real-world streaming systems and the limitation of fully supervised learning models. Works by Lu et al. [9] have reviewed thoroughly techniques and methods to achieve learning with ongoing concept drift in streaming data systems.

The use of deep learning in streaming data classification has been explored in works like Mohammadi et al. [10], focusing on architectures and training strategies suitable for incremental learning to tackle real-time data streams at the era of Internet of Things (IoT). Neumeyer et al. [11] focused on the scalability challenge, presenting distributed computing approach to manage large-scale streaming data efficiently.

Integrating streaming data classification with emerging technologies has revealed new research avenues. Kong et al. [12] provides a comprehensive survey of the role of edge computing in the context of the Internet of Everything (IoE).

An additional important aspect in the field of steaming data is benchmarking, with researchers aiming for standardized evaluation protocols to accurately assess and compare the performance of steaming data classification models. The work by Souza at el. [13] presents the use of dedicated repository that at their vision could act like a benchmark for researchers to compare their architecture for classification of streaming data.

Deep learning has advanced significantly with powerful computing resources, enabling the processing of large datasets without manual feature extraction. However, the continual influx of data presents challenges in training efficiency and storage. Dataset distillation emerges as a strategic response, condensing vast datasets into smaller, highly informative subsets. This approach differs from the core-set strategy of selecting informative samples directly. Instead, it focuses on synthesizing original datasets into a smaller number of samples optimized to encapsulate the original datasets' knowledge. Pioneered by Wang et al. [14], this methodology involves iteratively updating synthetic samples to ensure models trained on these samples excel with real data. This work has since spurred numerous subsequent studies, highlighting its impact and the broader shift towards more efficient data processing methods in deep learning [15, 16, 17].

In the context of these developments, our research contributes a novel image distillation technique for streaming data classification. This method, designed to efficiently refine and compress image streams, extracts relevant features while addressing the constraints of memory and computational power. Empirical testing has shown our approach outperforms established algorithms like HT, ARF, and RBC in accuracy.
We provide our code on GitHub to support and encourage further research in this area (https://github.com/RWAD777KH/Classification-of-Streaming-Data).

## 3   Methodology

In this methodology chapter, we present HT and ARF, both cornerstone algorithms for streaming data classification. Additionally, we examine the RBC technique, a method designed to tackle the challenges of working with vast data streams under limited memory constraints. Building on these established methods, we introduce our novel approach of incorporation data distillation technique into the continual learning process.

## 3.1 Hoeffding Trees and Adaptive Random Forest

At the core of the *Hoeffding Tree algorithm* (HT) is the use of Hoeffding bound to decide how many data instances need to be analyzed to make decisions at each node in the tree with a certain level of confidence. Hoeffding bound is a statistical measure that provides an upper limit on the probability that the sum of random variables deviates from its expected average value by more than a specific margin $\varepsilon$, regardless of the distribution of these variables. That is if $X_1, X_2, ..., X_n$ independent random variables, such that $a_i \leq X_i \leq b_i, i = 1, ..., n$, the Hoeffding bound is expressed as:

$$P\left(\left|\frac{1}{n}\sum_{i=1}^{n} X_i - E\left[\frac{1}{n}\sum_{i=1}^{n} X_i\right]\right| \geq \varepsilon\right) \leq 2\exp\left(-\frac{2n\varepsilon^2}{\sum_{i=1}^{n}(b_i - a_i)}\right) \quad (1)$$

The bound provides a way to statistically ensure that the decision made for splitting a node—based on the observed subset of streaming data—is nearly as good as the decision that would be made if all possible data had been observed. This statistical guarantee allows the algorithm to make split decisions with confidence, despite the inherent limitations in observing all data in a streaming environment.

The following steps detail the operational process of HT. As data streams the algorithm incrementally builds the decision tree by evaluating the data points that pass through each node, and for each node, it calculates the information gain for each possible feature to split on, using only the data that node had seen. Then it uses the Hoeffding bound to determine if the difference in the metric between the best feature and the second-best feature is significant enough to make a decision and if the bound is not exceeded it means that more data is needed to create an effective split feature. When a feature is selected for splitting, the tree grows by adding new nodes, and the algorithm repeats the process for each new node until maximum depth is reached or the improvement becomes negligible.

*Adaptive Random Forest* (ARF) enhances the conventional Random Forest approach by adjusting to evolving data streams. This adaptation makes ARF particularly suited for dynamic settings prone to concept drift, as it maintains an ensemble of decision trees like a traditional random forest. Each tree within this ensemble is developed on a data subset through feature and instance sampling, ensuring diversity among the trees. Typically utilizing Hoeffding Trees, ARF is capable of incremental growth from streaming data, allowing each tree to update with the arrival of new data without necessitating retraining from scratch— a process that could otherwise prove time-consuming. It continually assesses each tree's performance within the ensemble, replacing or modifying any that significantly drop in accuracy or fail to align with the current data distribution. For making predictions, ARF employs a weighted voting system based on each tree's accuracy, aggregating the ensemble's predictions to ensure that the most accurate trees have a greater influence on the outcome.

### 3.2 Reservoir Sampling Algorithm

Reservoir sampling is an algorithm for sampling a fixed number of items from a data stream of unknown size, ensuring each item has an equal chance of selection. It starts by filling a "reservoir" with the first items up to the desired sample size, then iteratively decides whether to replace them with subsequent items based on a probability that maintains representativeness. The main goal of random sampling is to ensure that every member of the population has an equal chance of being included in the chosen subset. This method helps achieve effective representatives, minimize bias, and with that acquire generalized results for the larger population.

We implemented the Stratified Reservoir Sampling technique in which the population or reservoir $R$ is divided into smaller groups, $R_c$, based on shared characteristics (classes). Mathematically, this can be represented as $R = \bigcup_{c=1}^{x} R_c$, where c indexes each class within R, and x is the total number of these classes. Then a simple random sample is taken from each stratum (group), and by this, we ensure representation across all key variables for all classes. Table 1 presents a pseudo code of Stratified Reservoir Sampling.

**Table 1.** Pseudo Code for Stratified Reservoir Sampling

| For each $R_c$ in R: | |
|---|---|
| **Initialize** $R_c$ | Array $R_c$ from 1 to k that contains the first k items of the input $\{x_1, \ldots, x_k\}$ |
| **In stream** $x_i \in \{x_{k+1}, \ldots, x_{k+B}\}$ | **For** each $x_i$: |
| B: stream size | Generate random number $j$ uniformly in range $\{1,\ldots,i\}$ |
| | **If** $j \in \{1, \ldots, k\}$: |
| | Set $R[j] := x_i$ |
| | **Else**: |
| | Discard $x_i$ |
| | Return updated $R_c$ |

### 3.3 Distillation Based Classification (DBC) Methodology

In our methodology, we present a novel strategy in the field of CSD by leveraging the data distillation technique. This technique focuses on the selective refinement and simplification of incoming data, aiming to boost both the accuracy and efficiency of the classification workflow.

Our process begins with the initial few batches of images, which are used to populate a uniform distillation reservoir $R_d$ of a predetermined size $S$. This reservoir serves as a dataset for fine tuning the ImageNet pretrained ResNet-34 model M and also initiates our distillation model $M_d$ ensuring both models are aligned from the outset. Upon the arrival of new batches of data $B$, we engage a sophisticated data distillation process $D_p$ that employs one of two distinct Convolutional Neural Networks (CNNs) - either a SimpleCNN or an IntermediateCNN. This step is critical for refining the reservoir content by selecting and updating it with more representative or informative data based on the new data.

The $D_p$ algorithm retrieves and updates the set of distilled images, taking as input the prior reservoir of distilled images $R_{d_{prev}}$ and a new batch of images. For each class, it individually processes these inputs using the CNN model $M_{CNN}$ in its training mode. This involves passing the inputs through the CNN model to obtain outputs for each input, denoted as $O_{m_{res}}$ and $O_{m_{batch}}$. Following this, the algorithm computes the loss between these outputs using Mean Square Error (MSE) loss function. It then backpropagates this loss and advances the optimizer, which results in updates to both the CNN model and the distilled images. Table 2 presents data distillation pseudo code.

**Table 2.** Pseudo Code of Distillation Process

| Inputs for the model | The prior reservoir of distilled images $R_{d_{prev}}$ |
|---|---|
| | New Batch of Images form the stream $B$ |
| | CNN model $M_{CNN}$ | (SimpleCNN or IntermediateCNN) |
| Process | **for** each *class in B*: |
| | $O_{m_{res}} = M_{CNN}(R_d^{class}{}_{prev})$ *[outputs for pervious distilled res]* |
| | $O_{m_{batch}} = M_{CNN}(B^{class})$ *[outputs for new batch B]* |
| | $Loss = MSE(O_{m_{res}}, O_{m_{batch}})$ |
| | Propagate $Loss$ |
| | Updated $M_{CNN}$ = **optimizer step()** |
| | **for** each class: |
| | $R_d^{class} = R_d^{class}{}_{prev} + learning\_rate * gradients\left(Loss\ w.r.t\ R_d^{class}{}_{prev}\right)$ |
| | $R_d = \sum_{i=0}^{nmber\ classes} R_d^i$ *[append all updated images for full updated reservoir of distilled images]* |
| Output | Updated $R_d$, Updated $M_{CNN}$ |

Subsequently, the primary model, M, is trained with the freshly distilled reservoir. This training is conducted for 10 epochs, or until we observe a degradation in loss for three consecutive times, allowing us to dynamically adjust to the distilled data's nuances (tiny differences). At predetermined intervals after every X batches, we perform validation checks on M against the current distillation reservoir, and preserving the model that demonstrates the highest validation accuracy.

This cycle repeats until the data stream—comprised of the training dataset—ceases, culminating in the application of the best-performing model $M_{best}$ to the test dataset for final inference.

Following at Table 3 is the pseudo-code for our algorithm methodology:

**Table 3.** Pseudo Code of DBC

| Initialize $R_d$ | Use first few batches to build $R_d$ uniformly with samples of all classes |
|---|---|
| $max_{acc}^{val} = 0$ | Initialize maximum validation accuracy to 0. |
| $R_d \rightarrow M$ | Train model $M$ with current distillation reservoir $R_d$ and log **acc & loss** |
| In stream $B_i \in \{x_1, ..., x_{B_{size}}\}$ | **for** each $B_i$:<br>　　$\{B_i \& R_d\} \rightarrow D_p$<br>　　Get updated $R_d$ from $D_p$<br>　　$R_d \rightarrow M$　*[Train model M with updated $R_d$]*<br>　　Log **t_acc & t_loss**<br>　　**if** Bi is multiple of **X:** *[each X batches we perform validation]*<br>　　　　Validate **M**<br>　　　　Log **v_acc & v_loss**<br>　　　　If v_acc > $max_{acc}^{val}$:<br>　　　　　　$max_{acc}^{val} = v\_acc$ *[update max validation accuracy]*<br>　　　　　　$M_{best} = M$　*[update best model]* |
| Test Best Model $M_{best}$ | Test $M_{best}$<br>Log **t_acc & t_loss** |

## 4　Experiments and Results

In this section, we present the experiments and results of the methods described in the previous section.

For all our experiments, we used the CIFAR-10 Dataset, a widely utilized resource in machine learning and computer vision research. This dataset comprises 60,000 color images, each of 32x32 resolution, evenly distributed across 10 categories, with 6,000 images per category. The dataset is divided into a training set of 50,000 images and a test set of 10,000 images.

### 4.1　Classification with HT and ARF

To adapt traditional streaming data algorithms HT and ARF for image classification, preprocessing is essential. This is achieved by using a ImageNet pretrained ResNet-34 which is adept at feature extraction from high-dimensional image data. ResNet-34 transforms raw images into a compressed, feature-rich representation suitable for these algorithms. This step enables the application of the algorithms to image streams.

In our study, we explored the capabilities of the Hoeffding Tree Classifier and the Adaptive Random Forest algorithm in classifying streaming data, utilizing the River library's default hyperparameters for both models. The approach for the HT involved processing data sequentially, updating the model with each new data point, showcasing the algorithm's strength in handling data streams one instance at a time. The preprocessing allowed us to adapt HT to image classification, achieving an accuracy of 53.2% with a training duration of 8000 seconds for one epoch.

Extending our investigation to the ARF, we maintained the library's standard configuration to test its performance under similar conditions. This ensemble method, designed for streaming data, comprises 10 trees without a maximum depth limit, employing information gain as the criterion for node splitting. Additional parameters included a split confidence of 0.01 and a minimum number of samples per leaf, with the learning rate set to 1. This configuration was selected to optimize the balance between computational efficiency and classification performance, tailored to the dynamic nature of streaming image data. The results obtained from processing the streaming data of image features and labels are compiled in Table 4 below.

**Table 4.** Results of Adaptive Random Forest

| # Trees | Validation accuracy | Test accuracy | Calculation Time |
|---|---|---|---|
| 10 | 31 | 31.28 | 60000 |
| 30 | 39.55 | 38.05 | 120000 |

In this scenario, the ARF algorithm processes each data point only once. However, its performance is still inferior to that of the Hoeffding Tree algorithm. This decreased effectiveness is likely due to the higher complexity of the ARF. Although ARF is a more powerful method, it may be more susceptible to overfitting in comparison to simpler models such as Hoeffding Trees. This is particularly true if the ARF's parameters are not meticulously optimized or if the dataset lacks sufficient variability to justify the use of such a complex ensemble approach. We also noticed that a training time of 12000 seconds per epoch is significantly longer than all other methods.

### 4.2   Reservoir Sampling based Classification (RBC)

In this experiment the model training and classification are done in a similar manner as described in the previous section (Table3). In optimizing the hyperparameters for the Reservoir Sampling algorithm, we explored various combinations of training batch sizes and reservoir sizes. The selection of an appropriate optimizer played a crucial role in navigating the algorithm's intricate landscape. After evaluating different optimizer types, we determined that the Adam optimizer was the most suitable choice. Additionally, the choice of loss function was pivotal in directing the algorithm's learning trajec-tory. Through experimentation, we found that employing the cross-entropy loss function resulted in the most optimal performance. Moreover, we fine-tuned the learning rate, a key determinant of convergence speed and stability. Ultimately, we settled on a final learning rate of 0.0001. Furthermore, we investigated the effects of implementing a learning rate decay mechanism to assess whether gradually decreasing the learning rate could enhance the learning process.

We started with batch size and reservoir size of 100 and performed throughout the training process 20 validation measurements on the continually updated model due to the extended runtime that the validation on every step would add. During every training and validation cycle, we recorded the accuracy and loss metrics. Additionally, we preserved the version of the model that achieved the highest validation accuracy at each iteration. Once all the training data had been processed, we conducted inferences on the test subset using the most effectively updated model.

We conducted similar experiments on this algorithm by utilizing various combinations of training batch sizes [100, 200, 500, 1000] and reservoir sizes [100, 200, 400, 1000, 5000]. The outcomes from the Reservoir Sampling Algorithm, for the best combination per train batch size, are compiled in Table 5 below:

**Table 5.** Results of RBC

| Training Batch Size | Validation accuracy | Test accuracy |
| --- | --- | --- |
| 100 | 70.48 | 69.53 |
| 200 | 71.06 | 70.9 |
| 500 | 68.33 | 67.81 |
| 1000 | 68.5 | 68.83 |

Based on these observations, we would suggest testing further combinations around a reservoir size of 100/200 with varying batch sizes since it shows a promising balance between training and test accuracies. Also, we observed that the overall time to train all the data in the best setup was 864 [second].

### 4.3 Distillation Based Classification (DBC)

We conducted an evaluation of the DBC following on previously described methodology. To optimize hyperparameters, we implemented a grid search strategy, fixing the reservoir size at 100 samples. This process involved assessing the impacts of various hyperparameters on performance, including the learning rate for model M, the learning rate for the distillation model $M_d$ the optimizer used for $M_d$, the type of CNN employed, and the number of epochs per batch.

During our experiment, we explored two optimization algorithms: Adam and Stochastic Gradient Descent (SGD). Adam differentiates itself by adjusting the learning rate individually for each parameter, leveraging estimates of the gradients' first and second mo-ments. This approach is designed to achieve quicker convergence, particularly beneficial in complex or non-convex optimization scenarios. Conversely, SGD employs a more straightforward mechanism, updating all parameters uniformly in a direction that minimizes the loss. This uniformity in learning rate application can render SGD's behavior more predictable and interpretable. Additionally, it's important to note that we did not apply any learning rate decay throughout this experiment.

In our framework, CNN is tasked with identifying disparities between the updated distilled reservoir $R_d$ and the incoming batch of images, processed in a stratified manner. SimpleCNN, with its simple design, aligns well with our preference for a system of minimal complexity, comprising a modest number of convolutional layers. On the other hand, IntermediateCNN seeks a middle ground, offering a design that is simple yet sufficiently complex. It incorporates more layers than SimpleCNN, aiming to enhance performance without significantly increasing computational demands. For a comprehensive overview of the CNN's implementation and a detailed exposition, refer to 22.

From our grid search analysis, it's discovered that IntermediateCNN outperforms SimpleCNN in terms of results without significantly extending the training duration. Additionally, we observed that the Adam optimizer shows a slight advantage over the SGD optimizer. A smaller learning rate for the Distillation model tends to yield better outcomes, and notably, when the Model's M learning rate is reduced, the accuracy between the training and validation phases becomes more aligned.

In conclusion, we've decided to proceed with further tests on two additional reservoir sizes [200, 500]. We're cautious not to exceed these sizes to avoid overly long computational times. The tests will utilize a combination of the IntermediateCNN and Adam Optimizer, exploring Model learning rates of [0.00/ 0.0001]. Additionally, with the lowest learning rate, we've decided to increase the epoch count to 20 when testing reservoir size of 500. The results from these experiments, focusing on streaming image data with distillation processing, are summarized in Table 6 below.

**Table 6.** Results of Classification of Streaming Images with distilled data

| Reservoir Size | Learning Rate | Epochs per Batch | Test accuracy |
| --- | --- | --- | --- |
| 100 | 0.001 | 10 | 62.53 |
| 200 | 0.0001 | 10 | 66.18 |
| 500 | 0.0001 | 20 | 73.1 |

## 5      Discussion and Conclusions

In our study, we aimed to innovate an algorithm that improves streaming image data classification within real-world computational and memory constraints. We compared our algorithm with existing models designed for streaming data, categorizing them into two groups: those that process data without storing images and those that retain images or their synthetic equivalents despite resource limitations. The performance, measured by accuracy and computational time for each model, is summarized in Table 7. Computational times refer to the average duration taken for training, validating, and testing over the span of 10 epochs. This comparison highlights our method's efficiency in managing the complexities of streaming data classification, optimizing performance under restricted system resources.

**Table 7.** Comparison of all methods: Accuracy [%] and Time [s]

| Method | HT | ARF | RBC | DBC |
|---|---|---|---|---|
| Accuracy | 53.2 | 39.55 | 70.9 | 73.1 |
| Time | 8000 | 10000 | 1000 | 1500 |

In our analysis of tree-based algorithms for streaming data, HT showed moderate effectiveness, balancing speed and accuracy but indicating room for improvement, especially with embedded categorical image data. The ARF, despite its ensemble learning strategy, recorded the lowest accuracy, underscoring the challenges of tuning such methods for streaming data where complex strategies don't always yield higher accuracy.

RBC, by contrast, was straightforward and relatively accurate without complex parameter tuning, though it didn't fully exploit the potential accuracy that typically achieved with no constrains theoretical model, especially under variable input distributions. Our novel approach, DBC, however, achieved superior accuracy, demonstrating its effectiveness in learning from streaming data and its potential to capture dataset complexities more accurately.

This comparison underscores the need for methodological adaptations to the unique demands of streaming data. While traditional methods like RBC provide insights, image distillation's conceptual novelty and alignment with theoretical accuracy expectations showcase its ability to effectively process streaming image data. Future work will delve into refining image distillation, assessing its performance across various data streams, and developing tailored CNN architectures to enhance its efficiency and application in real-world scenarios. Image distillation stands out as a promising direction for advancing classification models in streaming data, suggesting a pathway toward more sophisticated and resilient models.

**Disclosure of Interests.** The authors declare that they have no known competing financial interests or personal relationships that could have appeared to influence the work reported in this paper.